\documentclass[conference]{IEEEtran}
\usepackage{times}
\usepackage{epsfig}
\usepackage{graphicx}
\usepackage{amsmath}
\usepackage{amssymb}

\usepackage{algorithm}
\usepackage{algorithmicx}
\usepackage{algpseudocode}
\usepackage{subfigure}
\usepackage{hyperref}

\begin{document}

\title{ Core Sampling Framework for Pixel Classification}  

\author{\IEEEauthorblockN{Manohar Karki\IEEEauthorrefmark{1},
Robert DiBiano\IEEEauthorrefmark{2},
Saikat Basu\IEEEauthorrefmark{1},
Supratik Mukhopadhyay\IEEEauthorrefmark{1}
}
\IEEEauthorblockA{\IEEEauthorrefmark{1}Louisiana State University
Baton Rouge LA, 70803,USA}
\IEEEauthorblockA{\IEEEauthorrefmark{2} Autopredictive Coding LLC\\
Baton Rouge, LA, 70806, USA}

}

\maketitle

\begin{abstract}
The intermediate map responses of a Convolutional Neural Network (CNN) contain  information about   an image that can be used to extract contextual knowledge  about it. In this paper, we present a  core sampling framework that is able to use these activation maps from several layers as features to another neural network using transfer learning to provide an understanding of an input image. Our framework creates a representation that combines features  from the test  data and the contextual knowledge gained from the responses of a pretrained network, processes it and feeds it to a separate Deep Belief Network. We use this representation to extract more information from an image at  the pixel level, hence gaining understanding of the whole image. We  experimentally demonstrate the usefulness of  our framework using  a pretrained VGG-16 \cite{simonyan2014very} model to perform segmentation  on the BAERI dataset \cite{baeri_data} of Synthetic Aperture Radar(SAR) imagery and the CAMVID dataset \cite{BrostowSFC:ECCV08}. 

\end{abstract}

 \begin{figure*}
   \begin{center}
     \includegraphics[width=\textwidth]{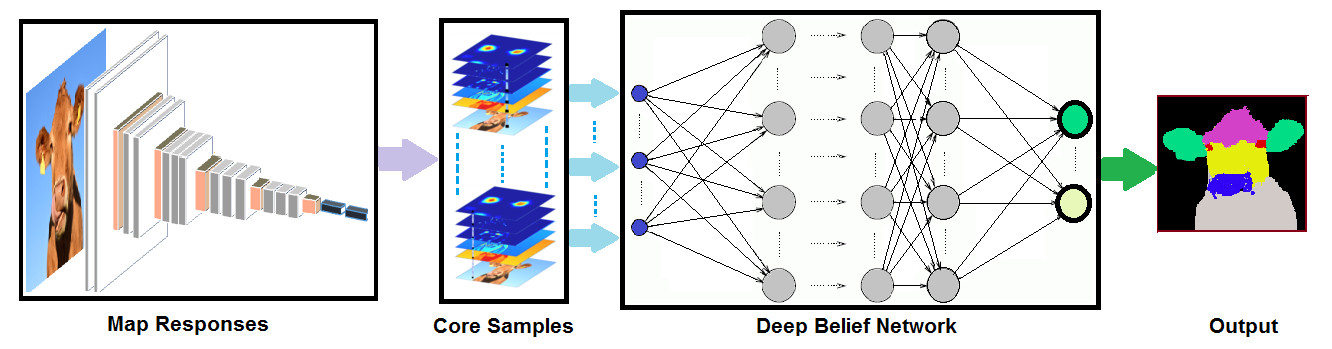}
    \caption{Core Sampling Framework. The map responses are generated by a pretrained network by passing images, which are processed as core samples and fed into a deep belief network. Output is prediction on each pixel of the test image, and can be different depending upon the task at hand. For example Segmentation (using classification), Colorization (using regression)  }
    \label{fig:framework}
   \end{center} 
   \end{figure*}
   
\section{Introduction}
Pixel-wise prediction/classification has a lot of applications \cite{hariharan2015hypercolumns} in scene understanding. It helps to generate a fine grain segmentation compared to existing techniques \cite{haralick1985image},\cite{barron2013volumetric},\cite{shi2000normalized} that segment at a coarser level. The rise of machine learning has led to novel and automatic segmentation techniques that require very little user input \cite{badrinarayanan2015segnet},\cite{long2015fully},\cite{girshick2016region}. Deep learning\cite{bengio2007greedy},\cite{ hinton2006fast}, \cite{lecun2015deep} in particular, has enabled this as it makes it possible to learn data representations without supervision.

\emph{Core sampling} has been used in engineering and science to understand the properties of natural materials\cite{lotter2003sampling}, climatic record from ice cores \cite{kotlyakov1985150} etc. Convolutional Neural Networks (CNNs), particularly useful for image understanding, work on parts of the input locally usually at different pyramidal levels.  In doing so, the network gathers  both low level and high level information about an input image. So, the lower layers encode the  pixels while the higer layers provide representation of  objects comprising of those pixels that eventually help in  understanding  the entire image.  Pixel wise classification and image understanding  can be improved  with the local and global information that are encoded in the different layers of a CNN; this information, stacked at different pyramidal levels, can be viewed as a \emph{core sample} that can enable better understanding of an image. A CNN consists of different layers that convolve the input image and/or parts of the image with filters. Training  a CNN involves  determining what these filters need to be in order to get the desired output for a given input. We can obtain map responses by feeding in test images through  these (pre)trained layers. Fig. \ref{fig:maps} shows some of the map responses that we get when we feed an input image of a cow through  the various layers of a CNN. 

In this paper, we present a core sampling framework that is able to use these activation maps from several layers as features to another neural network using transfer learning to provide an understanding of an input image. Our framework creates a representation that combines features  from the test  data and the contextual knowledge  gained from the responses of a pretrained network, processes it and feeds it to a separate Deep Belief Network. We use this representational model to extract more information from an image at  the pixel level, thereby gaining understanding of the whole image.

\begin{figure}
   \begin{center}
     \includegraphics[width=\columnwidth]{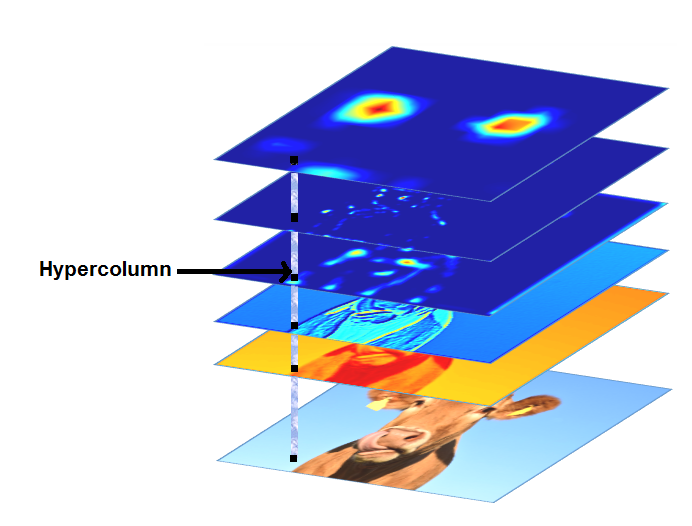}
    \caption{An example of a hypercolumn \cite{hariharan2015hypercolumns}. Each hypercolumn is defined as the vector of the values of the rescaled map responses at the exact same position in each of them.}
    \label{fig:hypercolumn}
   \end{center} 
   \end{figure}

 Image pyramids have  already been used to extract or learn more information from images  \cite{adelson1984pyramid}, \cite{burt1983laplacian} as well as for image segmentation  \cite{barron2013volumetric}. In a sense, the response from the layers of a CNN can also be viewed as the different pyramidal levels of the image viewed at different locations of that image. The notion of \emph{Hypercolumn}  introduced in \cite{hariharan2015hypercolumns}  builds upon this. The term Hypercolumn is used to describe a column of a group of map responses from the convolution layers of a network, aligned together such that each value in that column refers to the output of different layers from each of these maps, for an individual pixel as shown in Fig \ref{fig:hypercolumn}. In our framework, we accumulate such hypercolumns from each pixel of every training image and use that as training data to a deep belief network.
 
Transfer Learning  allows the use of the knowledge gained from solving one problem  to improve the solution to another  \cite{torrey2009transfer}. Our core sampling framework makes use of transfer learning, where  the core sample acquired from a previously trained network is used for training a second network. 
 This strategy helps increase the overall performance and speeds up training. The other advantage of using this approach is that it avoids the need to train  a very large network on a huge dataset. Often there is not enough training data for a particular task; this can be overcome by using the knowledge learned from a similar task \cite{pan2010survey} or a different task \cite{yosinski2014transferable}.


The two-stage architecture of our core sampling framework is given in Figure \ref{fig:framework}.  
We use  the VGG-16 \cite{simonyan2014very} model to bootstrap our framework.  This model has been trained on the ImageNet dataset \cite{krizhevsky2012imagenet} which consists of  more than a million training images and 1000 classes. The intermediate maps, that are generated for each pixel  during the testing phase when images are input to the above model, when stacked together and resized to a uniform size, form hypercolumns. 
A hypercolumn  for a pixel in an input image is a vector  with $k$ columns,  where $k$ is the number of  intermediate maps in  the VGG-16 model, with each component of the vector being a map. A hypercolumn does not preserve any spatial correlation between the constituent maps.  A \emph{core}  is a collection of hypercolumns, one per pixel in an input image.  A random sample drawn from a core is called \emph{a core sample}. Core samples generated from input images a fed to the second stage of our framework.

The second stage of our framework consists of a Deep Belief Network (DBN). Deep Belief Networks are  unsupervised deep learning models \cite{ hinton2006fast}. Since no spatial correlation among the maps is  preserved in the hypercolumns comprising the input  core samples, a CNN cannot be used in the second stage as the filters in a CNN presume spatial correlation between adjacent maps.  The DBN interprets the input core samples to provide an understanding of the original input image. 
%

This paper makes the following contributions. 
\begin{enumerate}
\item We present a novel core sampling framework that is able to use  activation maps from several layers  of a CNN as features to another neural network using transfer learning to provide an understanding of an input image. It creates a representation that combines features from the test  data and the contextual knowledge  gained from the responses of a pretrained network. This model can be used to  extract more information from an image at  the pixel level, thereby gaining understanding of the whole image.
\item We experimentally demonstrate the utility of the core sampling framework by showing its ability to automatically  segment    the  BAERI dataset \cite{baeri_data} of Synthetic Aperture Radar (SAR) imagery   and  the CAMVID dataset \cite{BrostowSFC:ECCV08}.  
\end{enumerate}
\section{Related Work}

CNNs have been used to extract information from images;  deep CNNs have enabled recognition of  objects in images with high accuracy without any human intervention \cite{ lecun1995learning}, \cite{krizhevsky2012imagenet},  \cite{bengio2009learning},  \cite{russakovsky2015imagenet}. There has been some research in using the information acquired from the intermediate layers of a CNN to solve  tasks such as classification, recognition, segmentation or a combination of these \cite{hariharan2015hypercolumns}, \cite{badrinarayanan2015segnet}, \cite{he2015deep}, \cite{ren2015faster}, \cite{long2015fully}. Image segmentation has been studied for decades.  Haralick and Shapiro \cite{haralick1985image} describe classical image segmentation techniques such as thresholding, multidimensional space clustering, region growing,  etc.  Comanciu and Meer \cite{comaniciu1997robust} use the mean shift algorithm to provide automatic segmentation, where human intervention is needed to choose the class of a segment. There are well known graph-based algorithms for image segmentation;   e.g., Normalized Cuts \cite{shi2000normalized}, where segmentation is achieved by measuring the similarity of graph partitions;  Graph Cuts \cite{ felzenszwalb2004efficient}, where a segmentation is defined as set of regions; this set of regions is repeatedly combined based on the similarity between neighboring regions.  Rother et. al. \cite{ rother2004grabcut} implement  iterative estimation on top of graph cuts algorithm \cite{boykov2001interactive} to define a boundary for the segmentation of objects where a user  selects the broader region. 
 \begin{figure*}
   \begin{center}
     \includegraphics[width=.65\textwidth]{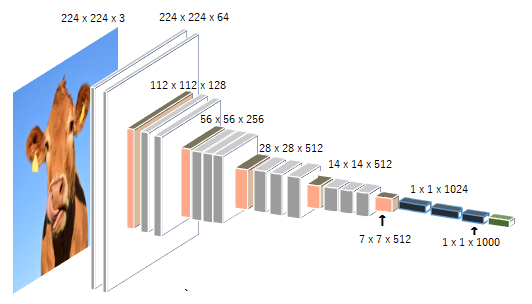}
    \caption{VGG-16 Archtiecture \cite{simonyan2014very} The architecture consists of multiple convolution layers, with max pooling layers in between and 3 fully connected layers followed by a softmax layer.}
    \label{fig:vggarch}
   \end{center} 
   \end{figure*}  
   
	Recent approaches to segmentation also include various ways to use a CNN to segment images. Girshick et. al. \cite{girshick2016region} use  Region-based CNNs (R-CNNs) where category-independent region proposals are defined during the pre-processing stage, that are input to a CNN to generate feature vectors.  A linear SVM (Support Vector Machine) is then used  to classify the regions. Unsupervised Sparse auto-encoders have been used in \cite{chen2014sar} on Synthetic Aperture Radar (SAR) data to classify different types of vehicles. They only deal with classification of images already segmented into smaller regions containing the objects. We deal with segmentation by classification at pixel level.  Ladicky et. al. \cite{ladicky2010and} use energy function for the conditional Random Fields (CRF) model  which they use to aggregate results from different recognizers. Zhang et. al. \cite{zhang2010semantic} recover dense depth map images and other information about the frames in video sequences such as height above ground, global and local parity, surface normal, etc. They use graph cut based optimization and decision forests to evaluate their features. The need of video sequences limits their application. Both the previous approaches need manual feature extraction. 
	Another approach is to use deconvolution layers after the convolution layers as a way to reconstruct segmented images as done in \cite{noh2015learning} and \cite{ badrinarayanan2015segnet}. SegNet \cite{ badrinarayanan2015segnet} uses an encoder architecture similar to  VGG-16. The decoder is constructed  by removing the fully connected layers and adding deconvolution layers. It is used to transform   low resolution maps to high resolution ones. The original  paper that proposed the notion of hypercolumns \cite{hariharan2015hypercolumns} also  uses the maps from the intermediate layers  of a CNN to segment and localize objects in images. They use a location specific approach;  in particular, they use $K$ x $K$ classifiers across different positions of the images where $K$ is a constant. A linear combination of these classifiers is used to classify each pixel. They also use bounding boxes of size 50 x 50 and try to predict the heat map to localize objects in an image. 

Deep Belief Networks (DBN) consist of multiple layers of stochastic, latent variables trained using an unsupervised learning algorithm followed by a supervised learning phase using feedforward backpropagation-based  Neural Networks.  In the unsupervised pre-training stage, each layer is trained using a Restricted Boltzmann Machine (RBM). Once trained, the weights of the DBN are used to initialize the corresponding weights of a Neural Network \cite{bengio2009learning}. A Neural Network initialized in this manner converges much faster than an otherwise uninitialized one.

In \cite{du2013unsupervised}, the authors  use transfer learning for segmentation of hyperspectral images by training on an unlabeled dataset and  then using transfer learning to improve separability  based on the learned knowledge.


\section{Core Sampling Framework for Pixel Classification}

We  use  hypercolumns  introduced in \cite{hariharan2015hypercolumns} as a data structure for representing  the layer outputs from a CNN. The first few layers are used for accurate localization of an object and the layers close to the output layer help to distinguish  between different objects. We use the pre-trained VGG-16 \cite{simonyan2014very} model  for bootstrapping  the core sampling framework.  Fig. \ref{fig:vggarch} shows the different layers of the  VGG-16 network. The architecture includes pooling layers between the convolution layers and fully connected layers at the end. This network is trained on the ImageNet dataset which contains a large variety of objects. This makes it a perfect model to construct a framework that  works for a variety of datasets \cite{yosinski2014transferable}. Our framework uses the intermediate maps that are acquired during the testing phase when images are input to this model. The individual pixels of the intermediate maps, when stacked together and resized to a uniform size, form hypercolumns. 
Each pixel's value, combined with the map values produced using the pretrained model, is used as a data point. The map values are thus the features for their respective pixels. While using the maps from just 5 convolutional layers of the pretrained model, the number of maps per image is already around 1500. As the size of the core sampling data (processed output data from the pretrained network, described in \ref{sssec:coresample}) gets large, we came up with two ways to handle the problem: a) saving the entire data into  the hard disk,  b) using the data from a randomly sampled subset of pixels to train the DBN in the second stage.

 \begin{figure*}
   \begin{center}
    \begin{tabular}{cccccc}
     \includegraphics[width=20mm]{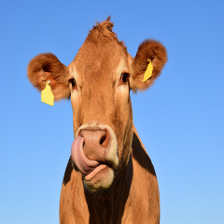} &
     \includegraphics[width=20mm]{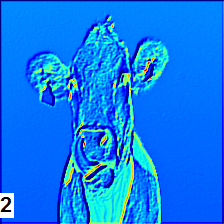} &
     \includegraphics[width=20mm]{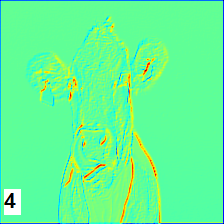} &
     \includegraphics[width=20mm]{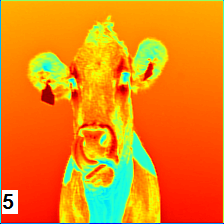} &
     \includegraphics[width=20mm]{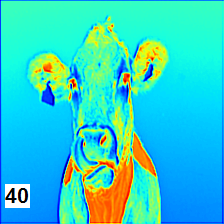} &
     \includegraphics[width=20mm]{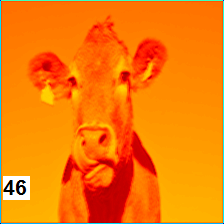} \\
     \includegraphics[width=20mm]{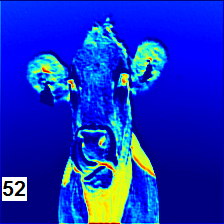} &
     \includegraphics[width=20mm]{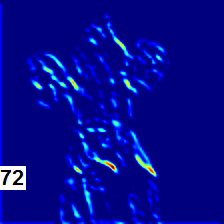}& 
     \includegraphics[width=20mm]{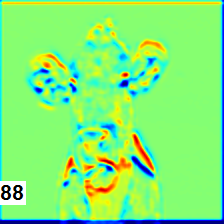} &
     \includegraphics[width=20mm]{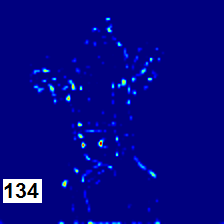} &
     \includegraphics[width=20mm]{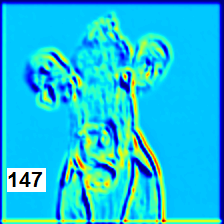} &
     \includegraphics[width=20mm]{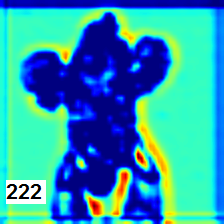} \\
      \includegraphics[width=20mm]{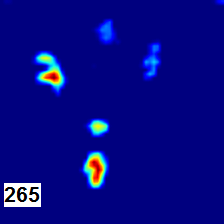} &
     \includegraphics[width=20mm]{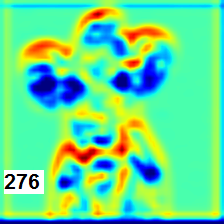}& 
     \includegraphics[width=20mm]{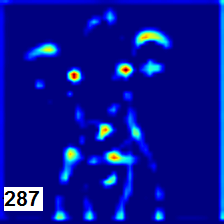} &
     \includegraphics[width=20mm]{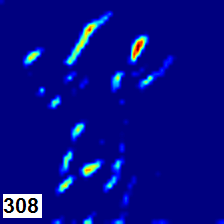} &
     \includegraphics[width=20mm]{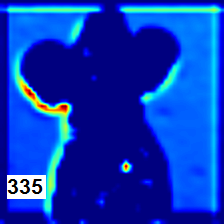} &
     \includegraphics[width=20mm]{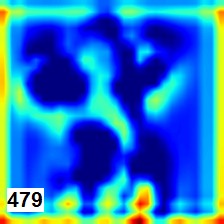} \\
      \includegraphics[width=20mm]{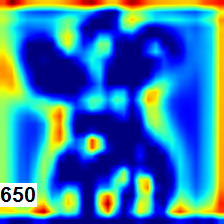} &
     \includegraphics[width=20mm]{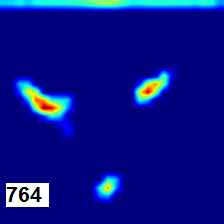}& 
     \includegraphics[width=20mm]{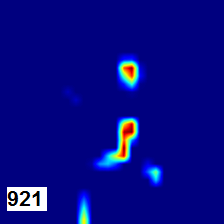} &
     \includegraphics[width=20mm]{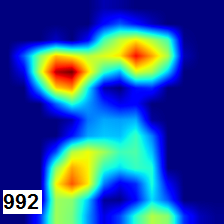} &
     \includegraphics[width=20mm]{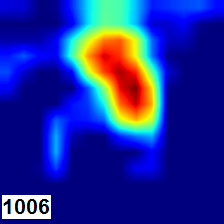} &
     \includegraphics[width=20mm]{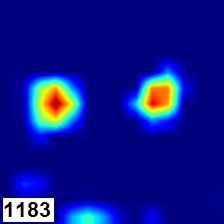} 
     \\
    \end{tabular}
    \caption{Response Maps resized to original image size. Higher number indicates maps from deeper layers.}
    \label{fig:maps}
   \end{center} 
   \end{figure*}

 \subsection{Preprocessing and Data Augmentation}
 The input to the pre-trained VGG-16 model needs to be of the size 224 by 224. The BAERI dataset (see Section \ref{sssec:dataset}) consists of raw images at inconsistent intensity levels and variable image sizes. Resizing the images would create images that are at different scales. So we added padding around smaller images to create 224 by 224 images. Because we did not resize them, the scale information remained  intact. For the same reason, we created sub-images (tiles) from larger images before extracting the map responses. These image tiles are created by using a sliding window of 224 by 224 and a smaller stride size. As before, there is no resizing of images; hence   there is no need for scale normalization. We also generated more data by varying contrast to improve robustness   to images that the framework  might not have seen and to create more training data for the next stage. The map responses, which are now used as features, are individually normalized and the same normalization parameters are used for the corresponding features during testing. 
All the images in the CAMVID dataset are of the same size (480 x 360) and are at the same scale,  being a standard dataset; hence not much preprocessing or data augmentation needs to be done. 

 \subsection{Response Maps}
 The layers of a deep neural network  learn different features at different layers or combinations of layers. The first layer of the network  learns features that are similar to Gabor Features or Color blobs \cite{ yosinski2014transferable}. The deeper layers help to discriminate objects and parts of objects while losing spatial and local information \cite{ zeiler2014visualizing}. Hence, the combination of maps at different layers helps to capture the spatial as well as the discriminative features. In \cite{ zeiler2014visualizing},  it is pointed out that removing the fully connected layers as features had resulted in very little increase in the error rate. Since the response from the fully connected layers is a vector (either of size  1024 or 1000)  resizing it would  drastically increase the size of  the data without  any significant increase in performance. We can see from Fig. \ref{fig:maps} that the deeper maps  extract more and more abstract features.  For example, the 1183\textsuperscript{rd} map  identifies the eyes/ears of the cow whereas the first few maps  detect the edges of the image. The deeper maps, however, lose the detailed spatial information about the objects.

\subsection{Core Sample: Intermediate Data Representation}\label{sssec:coresample}
Since, we use a pre-trained model to extract the maps,  we normalize the images by subtracting constants from the  R, G and B values (the same procedure that was used while training that model). From each image we then acquire  the map responses from each layer. Most of the map responses are n x n shapes (n $\in 2^i$, i = positive integer). Each of these map responses of various sizes are then resized to the size of the original image  using  bilinear interpolation. These map responses are then stacked along with original input image. From this point onwards, each pixel is a distinct data point with the map response values as its features. We are expecting a data point to have a single label value, when we train on this  data in the second stage  of our framework.  We define a \emph{core} as a collection of hypercolumns,  one per pixel for an input image. \emph{Core samples} are random samples drawn from a core. We feed the core samples to the second stage of our framework.  

\subsection{Stage 2: Pixel Prediction}
We use unsupervised pretraining using RBMs  followed by supervised learning using  DBNs  for the  final pixel-wise prediction. Our framework is flexible; depending upon the task at hand, different types of output layers can be used. We have implemented two different types of layers: a regression layer that implements linear regression and uses the mean-squared error as the loss function and a logistic regression  layer that can classify pixels using negative log likelihood as the loss function.  For example, if two pixels have red (R) values 0.5 and 0.6 they are more similar than if they were 0.1 and 0.6. The loss function that we typically use on regression problems is the mean squared error:
 \begin{equation}
\frac{1}{n}\sum_{i=1}^{n}(\hat{y}^{(i)}-y^{(i)})^2,
\end{equation}
 where $\hat{y}$ is vector of predicted values and $y$ is vector  of actual values for  $n$ observations. On most neural networks that classify rather than perform regression, the distance between any two labels would be the same. 
In such  cases, the  likelihood ($\mathcal{L}$) and loss ($l$) \cite{deeplearning.net} functions  are given by:

\begin{equation}\label{eqn1}
\begin{split}
\mathcal{L} (\theta=\{W,b\},\mathcal{D})=\sum_{i=0}^{\left | \mathcal{D} \right |} log(P(Y=y^{(i)}|x^{(i)},W,b))\\
l(\theta=\{W,b\},\mathcal{D})=-\mathcal{L}(\theta=\{W,b\},\mathcal{D}),
\end{split}
\end{equation}
where $W, b$ are weights and biases respectively and $\mathcal{D}$ is the dataset. Given an input, the weights matrix, and  a bias vector,   it outputs the likelihood that the input $x^i$ belongs to a certain class $y^i$. Since this equation is based on probability values rather than  distance measure, it is more suitable for classification.


 \begin{algorithm}
  \caption{Core Sampling FrameWork}\label{algorithm1}
  \begin{algorithmic}[1]
\Function{Main} {$mode, params, path$}
\If{$(mode = training)$}
	\State $X,Y \gets \Call{CoreSample}{\textbf{True},  path}$ 
	\State $model \gets \Call{trainDBN}{params,X,Y}$ \Comment{Trains the DBN and returns a trained interpretive model}
	\State $\Call{SaveModel}{decision\_model, params.name}$
\Else
	\State $X \gets \Call{CoreSample}{\textbf{False}, path}$
	\State $model \gets \Call{LoadModel}{params.name}$
	\State $output \gets \Call{Predict}{model, X}$
\EndIf
\EndFunction
     \Function{CoreSample} {$isTrain, path$}
          \State  $model \gets \Call{LoadPretrainedModel}{name}$
	\State $images, labels \gets \Call{LoadImages}{path}$
     	\For{\texttt{ \textbf{each} instance \textbf{in} images}}
             \State$ instance \gets  \Call{preprocess}{instance}$
     	  \State $feature\_maps \gets  \Call{GetFeature}{instance}$  
          	 \For{\textbf{each} \texttt{map \textbf{in} feature\_maps}}
        		    \State $ upscaled \gets \Call{imresize}{map, (W, H), 'bilinear'}$
           	    \State $core \gets \Call{reshape}{upscaled, W  X  H}$ \Comment{converts upscaled to a vector of length W x H}
           	    	\State $X \gets  \Call{concatenate}{X, core}$
	\EndFor
      \EndFor
      \State $X \gets \Call{Normalize}{X}$
	\If{$(isTrain)$}
		\State $Y=\Call{CreateTargets}{labels}$\Comment{Uses label images to create targets}
		\State \textbf{return} $X,Y$
	\Else
      		\State \textbf{return} $X$
	\EndIf
    \EndFunction

  \end{algorithmic}
      \end{algorithm}
      
      \subsection{Core Sampling Algorithm}
 Algorithm \ref{algorithm1} implements the core sampling framework. There are two modes of running the framework: training (Line 2) and testing (Line 6). In both cases, we generate core samples from images in a folder and normalize them before  training or testing  begins. The function CoreSample() loads the pretrained model and the images and iterates through the images. The aformentioned preprocessing/normalizing of the images is done and then each image is fed into the model. GetFeature() extracts the feature maps and then each map is upscaled to a uniform size of $W \times H$ using bilinear interpolation where $W$ and $H$ are the width and height of the input image. Then we concatenate all the cores into a single array and normalize them. The normalization parameters for each feature map are separately maintained.
During training, an array of target vectors is created by using the labeled images. Normalized samples from the cores and the labels are used as training data for the Deep Belief Network and the trained model is saved to hard disk. 
During testing, we load the model that we trained using our interpretive network (DBN) feed the core data in and finally make our prediction.

\subsection{ Datasets}\label{sssec:dataset}
 The CAMVID dataset \cite{BrostowSFC:ECCV08}, \cite{BrostowFC:PRL2008} consists of 32 semantic classes of objects out of which, like most of the other approaches  \cite{ladicky2010and}, \cite{zhang2010semantic}, we evaluate our algorithm on the 11 major classes and 1 class that includes the rest. These classes are Building, Tree, Sky, Car, Sign-Symbol, Road, Pedestrian, Fence, Column-Pole, Side-walk and Bicyclist. The training set includes input images, that are regular three channel color images and the targets are segmented single channel images. The images are from a few videos taken from inside a car driven on streets. These  consist of labeled images with   367 training, 101 validation and 233 test images of consistent sizes at the same scale.

The BAERI dataset \cite{baeri_data} that we introduce here consists of imagery  collected from a Synthetic Aperture Radar (SAR). Both the input and output are treated as single channel images. The input single channel image consists of SAR values and the labels are the ground truth values at each of the pixels. Labels were obtained by morphological image processing techniques for noise removal, i.e., opening and closing. The erosion and dilation on the images fail to remove all the noise in the ground truth images. Therefore, there are certain areas in the ground truth images that still contain some noise with incorrect labels. As, we are classifying each pixel separately, the noise must be taken into account during training.  The values in the training images were in the range between -40 and 25. In the BAERI dataset, the pixel classes are those belonging to  the ship class and the rest. There are only 55 images of variable sizes available in this dataset with the total size of 68 megabytes.
 \begin{figure*}
   \begin{center}
    \begin{tabular}{cccccc}
     \includegraphics[width=30mm]{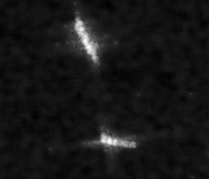} &
     \includegraphics[width=30mm]{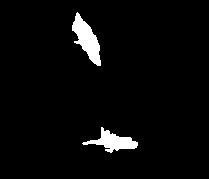} &
     \includegraphics[width=30mm]{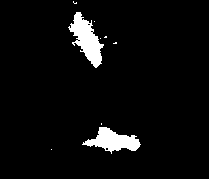} &
      \includegraphics[width=30mm]{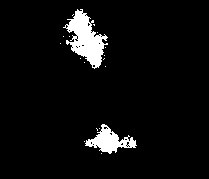} & \\
     \includegraphics[width=30mm]{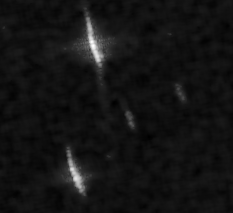} &
     \includegraphics[width=30mm]{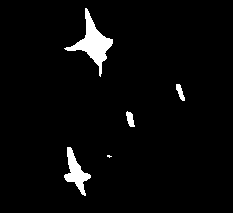}& 
     \includegraphics[width=30mm]{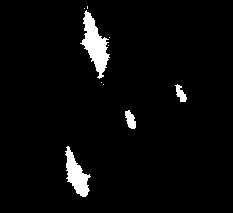} &
     \includegraphics[width=30mm]{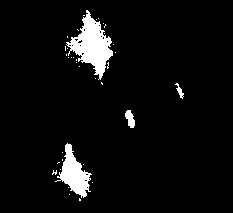} &\\
      \includegraphics[width=30mm]{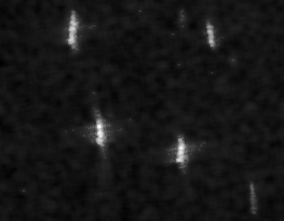} &
     \includegraphics[width=30mm]{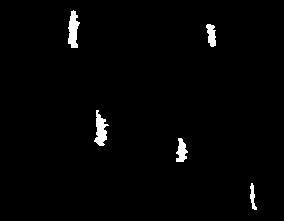} &
     \includegraphics[width=30mm]{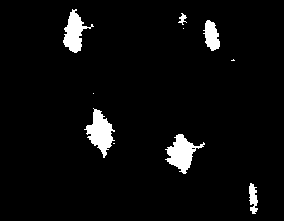} &
      \includegraphics[width=30mm]{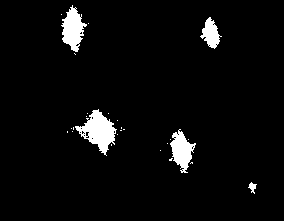} 
      \\
    \end{tabular}
    \caption{Results on the BAERI dataset. The images from left to right: a) Original Image b) Ground Truth c) Our Alogrithm d) SegNet \cite{badrinarayanan2015segnet}}
    \label{fig:results_baeri}
   \end{center} 
   \end{figure*}
   
    \begin{figure*}
   \begin{center}
    \begin{tabular}{cccccc}
     \includegraphics[width=30mm,height=30mm]{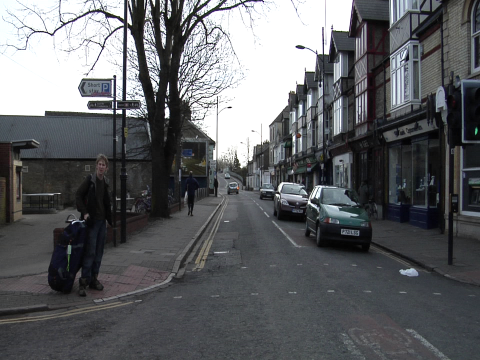}&
     \includegraphics[width=30mm,height=30mm]{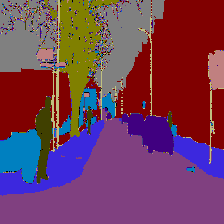} &
     \includegraphics[width=30mm,height=30mm]{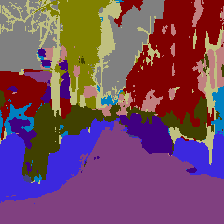} &
     \includegraphics[width=30mm,height=30mm]{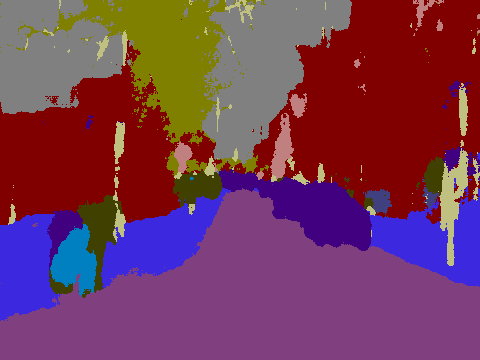} &\\
     \includegraphics[width=30mm,height=30mm]{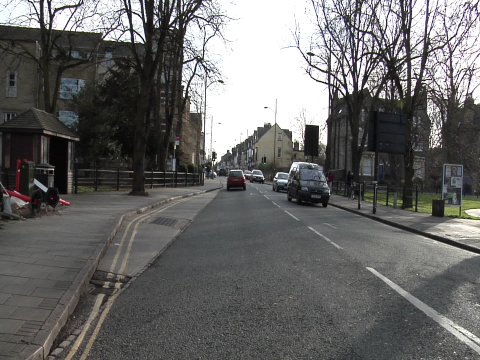} &
     \includegraphics[width=30mm,height=30mm]{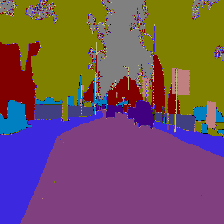} &
     \includegraphics[width=30mm,height=30mm]{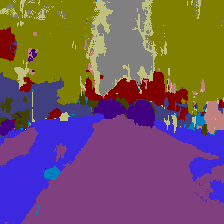} &
     \includegraphics[width=30mm,height=30mm]{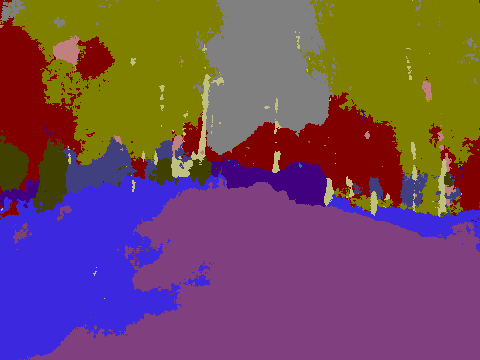} &\\
   \includegraphics[width=30mm,height=30mm]{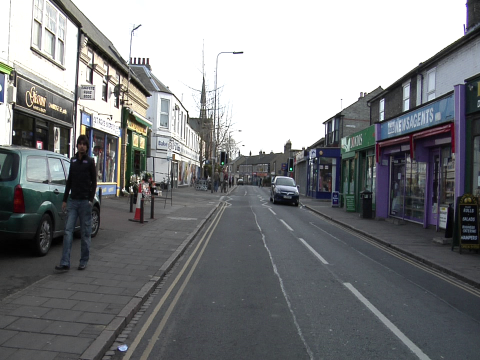} &
     \includegraphics[width=30mm,height=30mm]{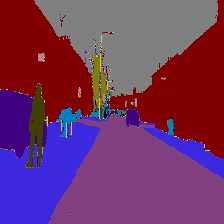} &
     \includegraphics[width=30mm,height=30mm]{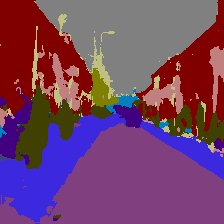} &
     \includegraphics[width=30mm,height=30mm]{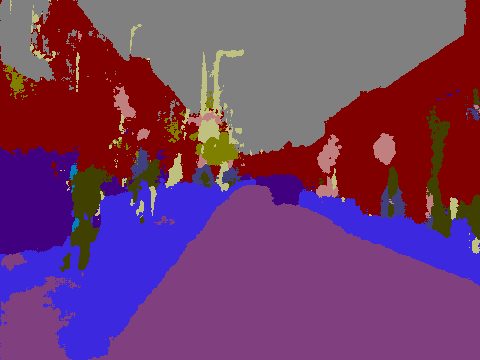} &\\
      \includegraphics[width=30mm,height=30mm]{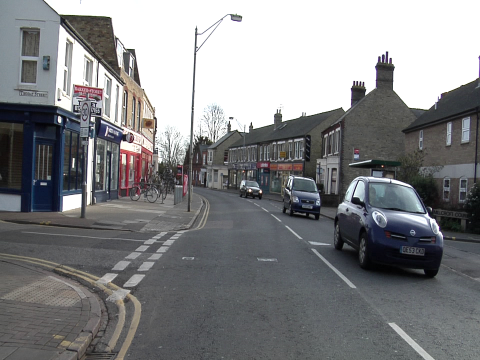} &
     \includegraphics[width=30mm,height=30mm]{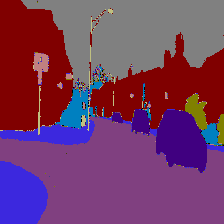} &
     \includegraphics[width=30mm,height=30mm]{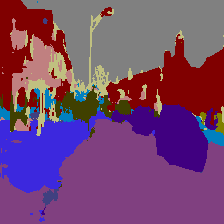} &
     \includegraphics[width=30mm,height=30mm]{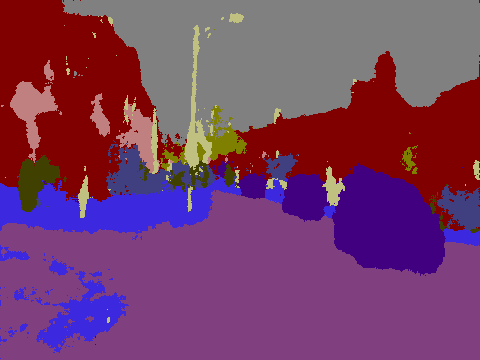} &\\
      \includegraphics[width=30mm,height=30mm]{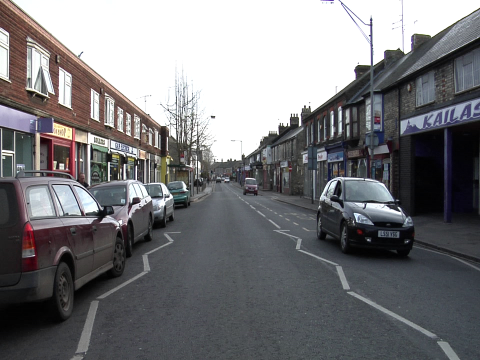} &
     \includegraphics[width=30mm,height=30mm]{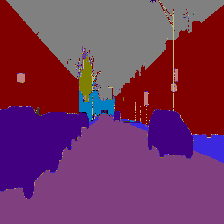} &
     \includegraphics[width=30mm,height=30mm]{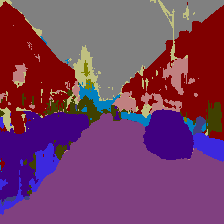} &
     \includegraphics[width=30mm,height=30mm]{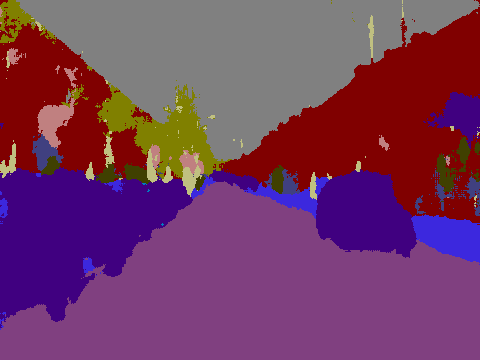}\\
    \end{tabular}
    \caption{Results on the CAMVID dataset. The images from left to right: a) Original Image b) Ground Truth c) Our Alogrithm d) SegNet \cite{badrinarayanan2015segnet}}
    \label{fig:results_camvid}
   \end{center} 
      \end{figure*}
\section{Experimental Results and Discussion}
Since we only deal with classification in this paper, we use  negative log likelihood as our cost function. We trained the DBN in the second stage  by first performing  a pre-training step with persistent chain contrastive divergence (P-CD) and then fine tuning the network using a deep  feedforward neural network trained by backpropagation. Keeping the pretrained CNN and the  DBN interpreting the core samples distinct makes it easier to preprocess the data and also to normalize the testing set with respect to the training data. We used L1 and L2 norms for regularization and implemented dropout on the hidden layers. The input to the DBN are the core samples that we described above. The map responses from each layer  of the CNN are normalized using standard feature scaling. The unsupervised training helps us to cluster the features together further, and helps to converge the training faster. We used the theano \cite{2016arXiv160502688short} deep learning library and an Intel i7 six core  server with  TITAN X GPU for our experiments.

Between the CAMVID dataset and the BAERI one,   there are more objects in the CAMVID dataset that are also present in the ImageNet dataset. On the other hand, the BAERI dataset is quite different from ImageNet because it contains SAR data  not present in ImageNet. As a result of transfer learning, the knowledge acquired from ImageNet based  the wide variety of features abstracted at various levels by the  pretrained VGG-16 network prevented the sparsity of the BAERI dataset from creating any problem  in training the DBN in the second stage. On the BAERI dataset, our frame work was able to slightly outperform SegNet (see Table \ref{table1}). The output images of SegNet had less blobs that could be classified as noise but it also missed a few of the smaller ships and had a larger area of pixels incorrectly classified as ships around the main clusters compared to our algorithm as can be seen on Fig. \ref{fig:results_baeri}. 

\begin{table}[]
\centering
\caption{Results on BAERI Dataset}
\label{table1}
\begin{tabular}{c|cc}

  Metric                   & Our Method & SegNet \\ 
    & & \cite{badrinarayanan2015segnet} \\ \hline
Accuracy (\%)            & \textbf{99.24} & 98.08                                                                              \\ 
Mean Squared Error (MSE) & \textbf{.0115} & .0142                                                                              \\ 
\end{tabular}
\end{table}

  On the CAMVID dataset, the core sampling framework  outperformed  both \cite{ladicky2010and}, \cite{zhang2010semantic} on 10 of the 11 classes in terms of accuracy and had a better per class accuracy (see Table \ref{table2}). Both \cite{ladicky2010and}, \cite{zhang2010semantic} and our framework were trained on 367 labeled training images. 
As can be seen from Table \ref{table2},  our framework could not match the performance of  SegNet  on the CAMVID dataset in terms of accuracy except for the Sky, Column-Pole, and Bicyclist classes where it outperformed SegNet. However, while our framework was trained on 367 labeled images, SegNet was trained on 3500 labeled images. 

\begin{table}[]
\centering
\caption{Results on CAMVID Dataset}
\label{table2}
\begin{tabular}{ccccc}
 Classes & Our 						& Boosting								& Dense									& SegNet\\
   	   & Method   					& (CRF + 								&Depth							 & \cite{badrinarayanan2015segnet} \\
  	   &							&Detectors)								&Maps 										&\\
  	   &							& \cite{ladicky2010and}  						&\cite{zhang2010semantic} &  \\ \hline
     & \multicolumn{3}{c}{367 Training Images} & 3.5 K Images                                       \\ \hline
Building              & 71.0                                                  & 81.5                                                                                              & 85.3                                                                                        & \textbf{89.6}                                                                      \\
Tree                  & 62.6                                                  & 76.6                                                                                              & 57.3                                                                                        & \textbf{83.4}                                                                      \\
Sky                   & \textbf{96.8}                                         & 96.2                                                                                              & 95.4                                                                                        & 96.1                                                                               \\
Car                   & 72.2                                                  & 78.7                                                                                              & 69.2                                                                                        & \textbf{87.7}                                                                      \\
Sign-Symbol           & 52.3                                                  & 40.2                                                                                              & 46.5                                                                                        & \textbf{52.7}                                                                      \\
Road                  & 80.4                                                  & 93.9                                                                                              & \textbf{98.5}                                                                               & 96.4                                                                               \\
Pedestrian            & 56.4                                                  & 43.0                                                                                              & 23.8                                                                                        & \textbf{62.2}                                                                      \\
Fence                 & 48.1                                                  & 47.6                                                                                              & 44.3                                                                                        & \textbf{53.5}                                                                      \\
Column-Pole           & \textbf{39.5}                                         & 14.3                                                                                              & 22.0                                                                                        & 32.1                                                                               \\
Sidewalk              & 77.3                                                  & 81.5                                                                                              & 38.1                                                                                        & \textbf{93.3}                                                                      \\
Bicyclist             & \textbf{38.5}                                         & 33.9                                                                                              & 28.7                                                                                        & 36.5                                                                               \\ \hline
\textbf{Class Avg.}   & 63.2                                                  & 62.5                                                                                              & 55.4                                                                                        & \textbf{71.2}                                                                      \\
\textbf{Global Avg.}  & 80.0                                                  & 83.8                                                                                              & 82.1                                                                                        & \textbf{90.4}                                                                     
\end{tabular}
\end{table}

\section{Conclusions}
We presented a  core sampling framework that is able to use the activation maps from several layers of a CNN as features to another neural network using transfer learning to provide an understanding of an input image. We  experimentally demonstrate the usefulness of  our framework using  a pretrained VGG-16 \cite{simonyan2014very} model to perform segmentation  on the BAERI dataset \cite{baeri_data} of Synthetic Aperture Radar(SAR) imagery and the CAMVID dataset\cite{BrostowSFC:ECCV08}. 
In the future, we intend to use the  core sampling framework to facilitate compression of images, texture synthesis, etc.

\bibliographystyle{IEEEtran}
\bibliography{egbib}

\end{document}